\documentclass[letterpaper, 10 pt, conference]{ieeeconf}  % Comment this line out if you need a4paper

\IEEEoverridecommandlockouts                              % This command is only needed if 
\usepackage{graphics} % for pdf, bitmapped graphics files
\usepackage{epsfig} % for postscript graphics files
\usepackage{amsmath} % assumes amsmath package installed
\usepackage{amssymb}  % assumes amsmath package installed
\usepackage{algorithm}
\usepackage{algpseudocode}
\usepackage{amsfonts}
\usepackage{mathtools}

\usepackage{times}
\usepackage{graphicx}
\usepackage{subfloat}
\usepackage{epstopdf}

\usepackage{multirow}
\usepackage{multicol}
\usepackage{slashbox}
\usepackage{adjustbox}
\usepackage{makecell}
\usepackage{cite}
\usepackage{here}
\usepackage{xcolor}

\definecolor{myred}{rgb}{1 0.5 0.5}

\makeatletter
\let\NAT@parse\undefined
\makeatother
\usepackage[linkcolor=red,citecolor=green,urlcolor=blue]{hyperref}
\usepackage[capitalise]{cleveref}

\title{\LARGE \bf
Robust Real-time RGB-D Visual Odometry in Dynamic Environments \\ via  Rigid Motion Model
}

\author{Sangil Lee, Clark Youngdong Son, and H. Jin Kim% <-this % stops a space
\thanks{Sangil Lee, Clark Youngdong Son, and H. Jin Kim are with the Department of Mechanical and Aerospace Engineering, Seoul National University, Seoul, 08826, Korea, Republic of {\tt\small \{sangil07, clark.y.d.son, hjinkim\}@snu.ac.kr}
}%
}

\begin{document}

\maketitle
\thispagestyle{empty}
\pagestyle{empty}

%%%%%%%%%%%%%%%%%%%%%%%%%%%%%%%%%%%%%%%%%%%%%%%%%%%%%%%%%%%%%%%%%%%%%%%%%%%%%%%%
\begin{abstract}

In the paper, we propose a robust real-time visual odometry in dynamic environments via rigid-motion model updated by scene flow. The proposed algorithm consists of spatial motion segmentation and temporal motion tracking. The spatial segmentation first generates several motion hypotheses by using a grid-based scene flow and clusters the extracted motion hypotheses, separating objects that move independently of one another. Further, we use a dual-mode motion model to consistently distinguish between the static and dynamic parts in the temporal motion tracking stage. Finally, the proposed algorithm estimates the pose of a camera by taking advantage of the region classified as static parts. In order to evaluate the performance of visual odometry under the existence of dynamic rigid objects, we use self-collected dataset containing RGB-D images and motion capture data for ground-truth. We compare our algorithm with state-of-the-art visual odometry algorithms. The validation results suggest that the proposed algorithm can estimate the pose of a camera robustly and accurately in dynamic environments.

\end{abstract}

%%%%%%%%%%%%%%%%%%%%%%%%%%%%%%%%%%%%%%%%%%%%%%%%%%%%%%%%%%%%%%%%%%%%%%%%%%%%%%%%
\section{INTRODUCTION}

Visual odometry is a fundamental process of recognizing the pose of the camera itself using video input\cite{nister2004visual,scaramuzza2011visual}. Although various visual odometry algorithms already show satisfactory performances in well-conditioned environments and well-defined datasets such as TUM\cite{sturm2012benchmark} and KITTI\cite{geiger2013vision}, most of them assume that the world the camera is looking at is stationary, thus making it possible to estimate the pose of the camera by virtue of the motion of the images taken. However, most real environments involve dynamic situations such as residential roads, crowded places, other robots for cooperation. Although some visual odometry algorithms \cite{dib2015robust} which utilize the principle of RANSAC can regard pixels whose motion is disparate as outliers, they have a limit that non-stationary objects should occupy small areas in the image plane. Thus, they cannot be employed to estimate the motion of a camera in dynamic environments including large non-stationary objects. 

In this paper, we aim for a real-time robust visual odometry by separating stationary parts from the image via motion model update. However, the existing motion segmentation algorithms have expensive computation loads\cite{sabzevari2014monocular} or constraints on the shape\cite{tsai2016video} or the number\cite{jung2014rigid} of objects. Here, we design a fast motion segmentation algorithm with an adequate performance applicable for real-time visual odometry algorithm. 

\begin{figure}[t]
\begin{center}
    \includegraphics[trim={0 1cm 0 0},width=8cm, height=7cm]{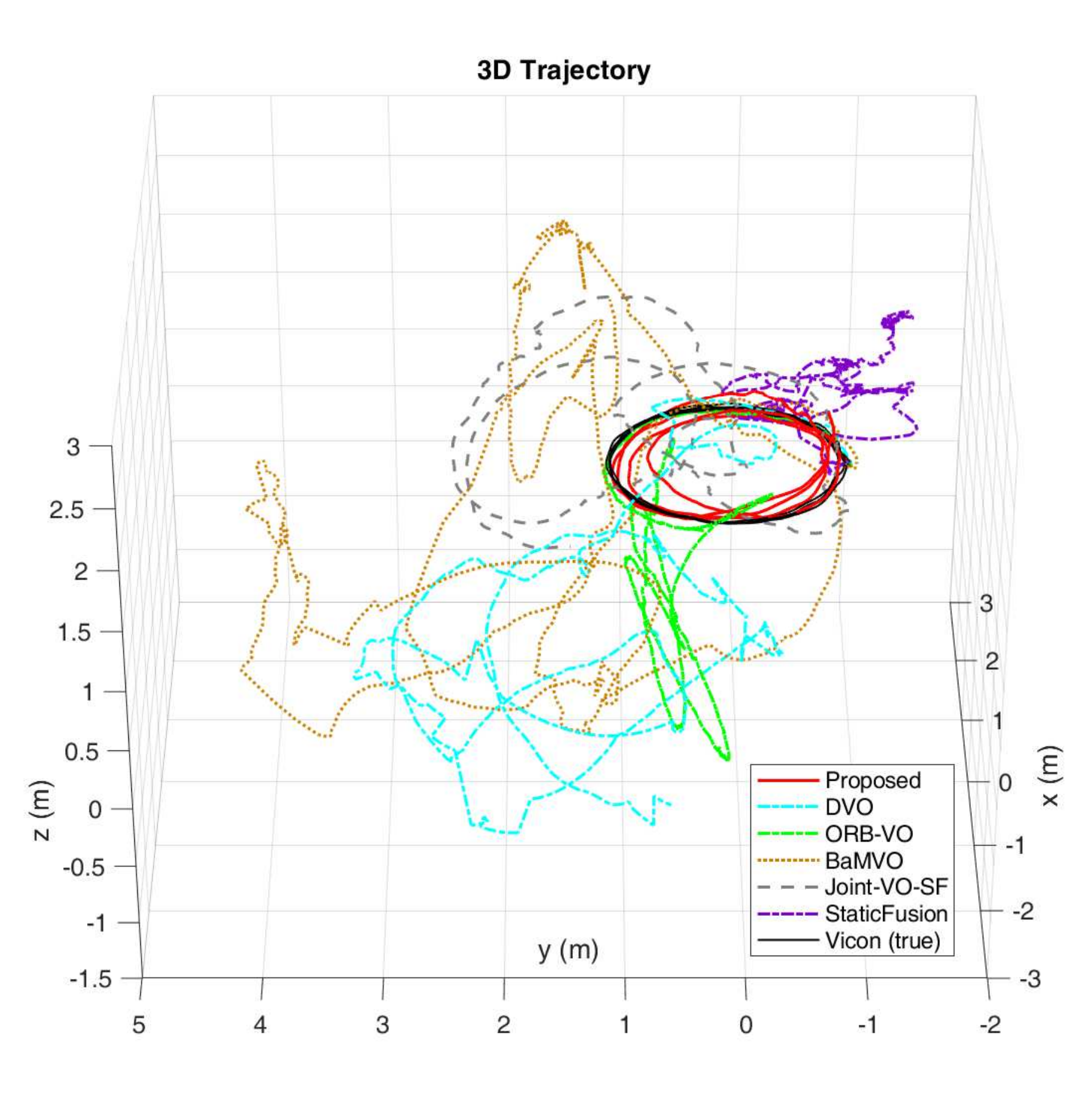}
    \caption{The 3D trajectory on the \texttt{uav-flight-circular} sequence.}
    \label{fig:dataset}
    \vspace{-8mm}
\end{center}
\end{figure}

In order to differentiate between non-stationary parts and stationary background, we utilize scene flow vectors which are distributed uniformly in the image. Particularly, we choose grid-based scene flow to take advantage of both dense and sparse methods; a dense flow that calculates temporary motions for all pixels provides a high resolution, meanwhile, a sparse flow that calculates temporary motions for distinctive features has a lighter computational load. Then, motion segmentation is performed to differentiate between non-stationary parts and stationary background with grid-based temporary motions, and the pose of a camera is estimated using static parts.

Overall, our algorithm can estimate the ego-motion robustly and accurately in highly dynamic environments while separating non-stationary parts from an image with no prior information such as the shape\cite{tsai2016video}, the number of dynamic objects\cite{jung2014rigid}, or the movement of objects\cite{kim2016effective}. \cref{fig:dataset} shows the performance of the proposed algorithm on the dataset collected from a multirotor flight. Moreover, the proposed algorithm shows significantly low runtime of average 19 ms at VGA resolution, thus it can be applied to real-time tasks.

\begin{figure*}[t]
\begin{center}
    \vspace{1.1mm}
    \includegraphics[width=.95\textwidth]{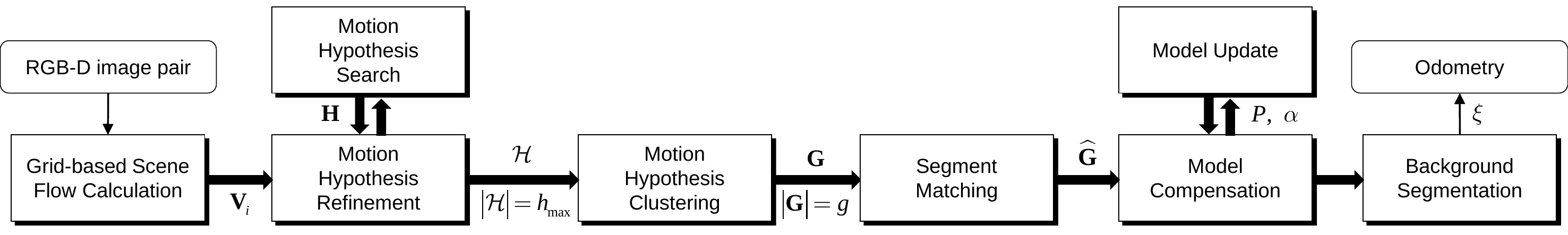}
    \caption{A schematic diagram of the proposed algorithm.}
    \vspace{-7mm}
    \label{fig:framework}
\end{center}
\end{figure*}

%-------------------------------------------------------------------------
\subsection{Related Work}

Most existing visual odometry algorithms \cite{steinbrucker2011real,kerl2013dense,forster2017svo,mur2017orb,engel2017direct} assume stationary environments. However, in real applications, there are often a number of non-stationary objects such as vehicles. To deal with such environments, some research has attempted to improve robustness against dynamic situations. Such efforts can be categorized into two types: dichotomous and model-based.

The first type of research has tried to exclude a region which has a different motion from the major movement in a similar way to RANSAC as mentioned above. A. Dib \textit{et al.}\cite{dib2015robust} utilizes RANSAC for direct visual odometry in dynamic environments. They minimize photometric shift over six random patches in an image, unlike the na\"ive direct visual odometry whose optimization process covers the whole image. In \cite{jung2008stereo}, they first categorize patches by depth value and subtract a category which has a quite different motion from a majority movement of the camera by comparing standard deviation of motion vectors, including or excluding that category. They still assume that the stationary background occupies a large area on the image so that the camera movement is estimated via the background.

The second type of research has exploited statistical models to classify motions belonging to independent objects. BaMVO \cite{kim2016effective} estimates background by choosing pixels whose depth does not change unusually. It is effective in situations where a dynamic object such as a pedestrian moves in parallel with the principal axis of a camera. However, the performance can degenerate when the dynamic object moves perpendicular to the principal axis while reducing the depth transition. Joint-VO-SF\cite{jaimez2017fast} formulates a minimization problem with 3D points from RGB-D images. They estimate camera pose and scene flow accurately, but the average runtime is 80 ms on an i7 multi-core CPU at QVGA resolution, which is slightly slow for real-time implementation with current on-board computers. StaticFusion\cite{scona2018staticfusion} is the enhanced version of Joint-VO-SF. It adds dense 3D modeling of only the static parts of the environment and reduces the overall drift though frame-to-model alignment.

Among the motion segmentation methods, H. Jung \textit{et al.}\cite{jung2014rigid} propose randomized voting to extract independent motions using epi-polar constraints with the average computational time of 300ms per frame. However,  the number of moving objects should be known. In \cite{moo2013detection}, MCD5.8ms has an advantage in that it detects a moving object using a dual-mode model with a low computational load while showing the execution time per frame of 5.8 ms. Since they calculate the homography using na\"ive visual odometry, however, the performance can deteriorate  when a moving object occupies more than half of an image. 

\subsection{Contributions}

In this paper, we focus on the environment where there are rigid moving objects. Furthermore, we propose a dichotomy-type algorithm to prevent the pose estimates from being polluted due to pixels with vague states between static and dynamic.

Our main contributions can be summarized as follows: 
\begin{enumerate}
	\item We propose a real-time robust visual odometry algorithm in dynamic environments. It estimates the motion of both stationary and non-stationary parts robustly.
	\item We design a dual-mode motion model for rigid motion segmentation. It distinguishes between the background and moving objects with no prior information.
\end{enumerate}

\section{BACKGROUND}\label{sec:backgrnd}

\subsection{Rigid Transformation}

The proposed algorithm is based upon the property that 3D motions of 3D points belonging to the same rigid object spatially have the same rigid motion temporally. We calculate the motion, $\mathbf{H}$, by the least-squares rigid transformation\cite{eggert1997estimating}. Moreover, for the evaluation of the motion, we can define the rigid transformation error as below:

\footnotesize{\begin{align}\label{eq:err}
	\mathbf{E}(\mathbf{H},\tilde{\mathbf{X}}^{(j)},\tilde{\mathbf{X}}^{(k)})\coloneqq\textrm{diag}( (\mathbf{H} \tilde{\mathbf{X}}^{(j)}\hspace{-.5mm}-\hspace{-.5mm}\tilde{\mathbf{X}}^{(k)})^T (\mathbf{H} \tilde{\mathbf{X}}^{(j)}\hspace{-.5mm}-\hspace{-.5mm}\tilde{\mathbf{X}}^{(k)}) ),
\end{align}}\normalsize
where $\tilde{\mathbf{X}}^{(l)} = [\tilde{\mathbf{x}}_1^{(l)}, \tilde{\mathbf{x}}_2^{(l)}, \ldots, \tilde{\mathbf{x}}_n^{(l)}]\in\mathbb{R}^{4 \times n}$ is a 3D point set, and $\tilde{\mathbf{x}}_*^{(l)} = [x, y, z, 1]^T\in\mathbb{R}^{4 \times 1}$ is a 3D point in the $l$-th frame represented by homogeneous coordinates.

\subsection{Advanced Grid-based Scene Flow} \label{sec:2:sceneflow}

In order to extract a 3D motion of pixels, we utilize Lucas-Kanade optical flow. For expanding the dimension of optical flow, we use depth changes from the adjacent depth images to obtain depth-directional flow. However, since there can be invalid depth pixels, we improve the quality of depth values through preprocessing. We fill pixels whose depth value is invalid with similar value in the vicinity in case of narrow holes and eliminate pixels whose depth value is abnormally high or low. Afterward,  we interpolate the depth of tracked points. 

\section{MOTION SPATIAL SEGMENTATION}\label{sec:mss}

\begin{figure}[t]
\begin{center}
    \vspace{2mm}
    \includegraphics[width=8cm,height=7cm]{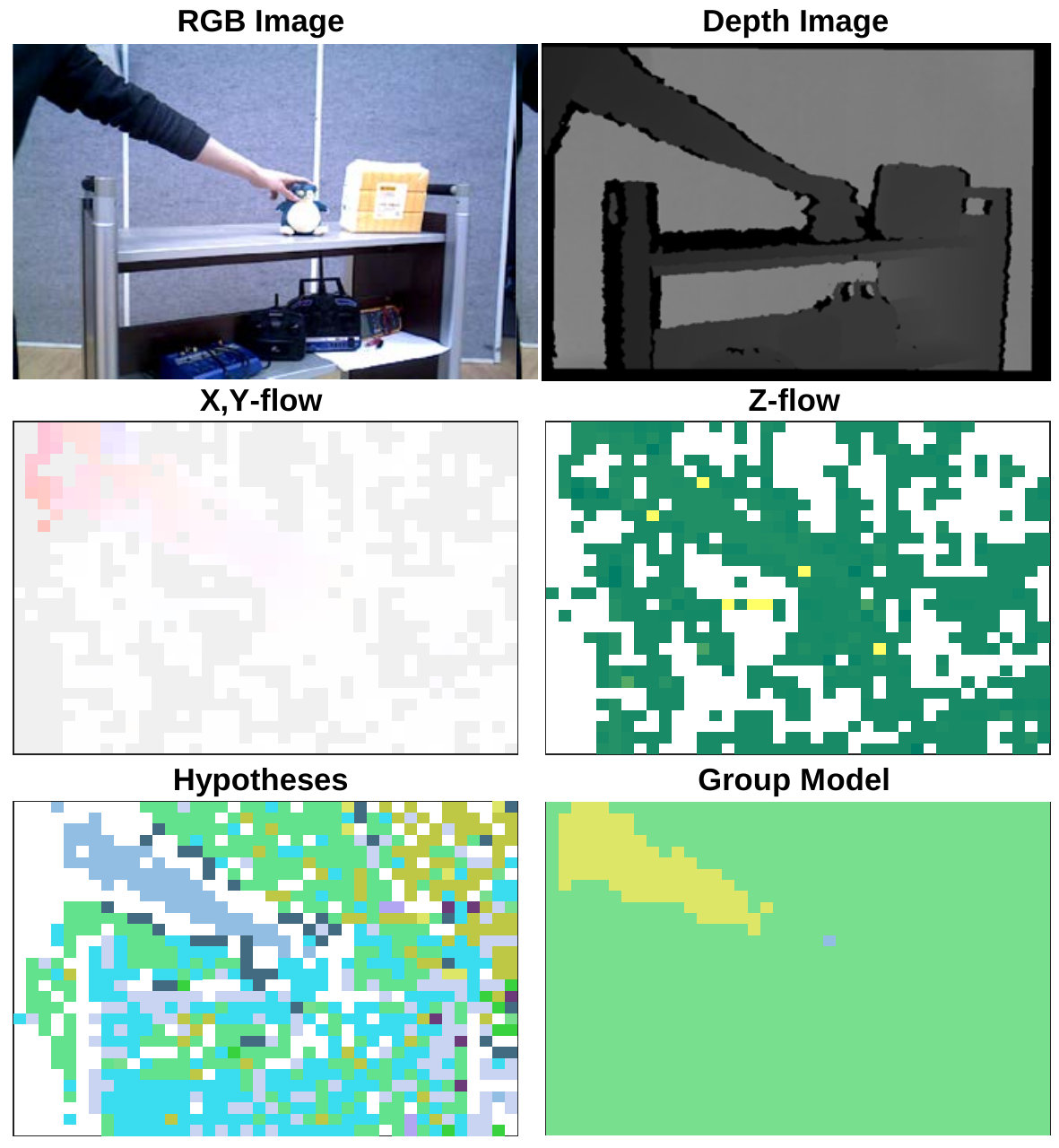}
    \caption{The description of the motion segmentation on the \texttt{place-items} sequence.}
    \label{fig:moseg}
    \vspace{-5mm}
\end{center}
\end{figure}

In the motion spatial segmentation procedure, we first fetch a total of $n$ grid-based scene flow vectors, $\mathbf{V}_i$, with grid cell size, $w_{\rm{grid}}$, and derive segments through their motions. In the procedure, the number of points $m$ and heuristic threshold $th_{\rm{inlier}}$ are used for generating motion hypothesis. To be specific, $m$ number of points are used to generate motion hypothesis, and the threshold value is used to decide whether motions of the $m$ points belong to the same movement. The motion spatial segmentation procedure is divided into three steps: motion hypothesis search, refinement, and clustering. Especially, the searching and the refining processes are executed iteratively until a total of $h_{\textrm{max}}$ hypotheses are found or the mean of entropies $\mathbf{S}$ is decreased enough and saturated. \cref{fig:moseg} describes the motion spatial segmentation process.

\subsection{Motion Hypothesis Search}\label{sec:MHS}

In the search process, a point is randomly selected by weighting with entropy, $\mathbf{S}$, whose  $i$-th value is defined as follows:
\begin{equation}\label{eq:weighting}
	S_i = 1-\min_{1\le h\le h_{\rm{max}}}{\exp(-\lambda \hspace{.5mm} \mathcal{E}_i - \delta)} , ~ i = 1 \ldots n,
\end{equation}
where
\small{\begin{align}
	\mathcal{E}_i = \left[ E_{i}(\mathbf{H}_1, \tilde{\mathbf{X}}^{(j)}, \tilde{\mathbf{X}}^{(k)}), \ldots, E_{i}(\mathbf{H}_{h_{\rm{max}}}, \tilde{\mathbf{X}}^{(j)}, \tilde{\mathbf{X}}^{(k)}) \right].
\end{align}}\normalsize
In the above equations, $E_{i}\left(\cdot\right)$ is the $i$-th element in the output of the $\mathbf{E}\left(\cdot\right)$ function defined in \cref{eq:err}. Also, $\lambda$ and $\delta$ are parameters for defining the characteristic of the entropy.

As will be discussed in more detail in the following section, the entropy measures how well the hypothesis is generated in the corresponding pixels. A pixel with the high entropy tends to belong to an inappropriate motion hypothesis, which results in a high probability to be chosen, consequently improving the convergence rate. Then we select $(m-1)$-points randomly near the chosen 1-point. Finally, the rigid motion hypothesis $\mathbf{H}$ can be estimated from the selected $m$-points\cite{eggert1997estimating}.

\subsection{Motion Hypothesis Refinement and Clustering}\label{sec:MHR}

Motion hypothesis refinement aims to calculate precise rigid transformation and find multiple regions with the same rigid motion. We first calculate rigid transformation error using \cref{eq:err} for all of the $n$ grid-based scene flow vectors. This process is designed to consider the cases where a part of a rigid object could appear in multiple regions on an image at the same time, in which case, we regard the multiple regions as belonging to the same object. Then we re-estimate a refined motion hypothesis using the increased $N$-points. For more details, please refer to \cite{lee2017real}.

In order to find distinct motions from the motion hypothesis set $\mathcal{H}$, we use the existing clustering algorithm, in particular, density-based spatial clustering of applications with noise (DBSCAN), since DBSCAN does not require the number of clusters and is robust to noise. In the process, hypotheses are refined as the algorithm reorganizes $h_{\textrm{max}}$ hypotheses into $g$ distinct motions, which is represented by vector $\mathbf{G} \in \mathbb{N}^{n \times 1}$ whose elements are distributed from 1 to $g$ in an natural number. Thus, a $\mathbf{G}^{(k)}$ is the segmentation results of the $k$-th frame at the motion spatial segmentation stage.

\section{MOTION SEGMENT TRACKING}\label{sec:mts}

In addition to distinguishing objects which move independently from one another in a frame, we propose a motion segment tracking algorithm to update the region of dynamic objects persistently. Since the above motion spatial segmentation algorithm runs in a similar way as the na\"ive segmentation technique based on two frames, it has no idea which is which at different times. Thus, we propose the motion segment tracking algorithm following the motion spatial segmentation. The following subsection provides a segment matching algorithm and a probabilistic approach based on the dual-mode simple Gaussian model \cite{moo2013detection}.

\subsection{Segment Matching}\label{sec:4:matching}

The segment matching algorithm calculates a kind of correlation coefficient and finds corresponding segment pairs $\hat{\mathcal{P}}$ between different frames. The correlation coefficient and segment pairs, $\mathcal{P}^{(k,l)}$, between the current $k$-th and the previous $l$-th frames are defined as follows:
\begin{align}
	C_{ij}^{(k,l)} &= \frac{N_{i,j}^{(k,l)}}{\sqrt{N_i^{(k)} N_j^{(l)}}},\\
    \mathcal{P}^{(k,l)} &= \left\lbrace\mathcal{P}^{(k,l)}_i\right\rbrace = \left\lbrace\left(\mathcal{P}^{(k,l)}_{i,\textrm{left}},\mathcal{P}^{(k,l)}_{i,\textrm{right}}\right)\right\rbrace \\\nonumber &= \left\lbrace\left(i, ~\underset{j}{\textrm{argmax}} C_{ij}^{(k,l)}\right)\right\rbrace,
\end{align}
where $N_i^{(k)}$ is the number of grid-based points belonging to the $i$-th segment in the $k$-th frame, and $N_{i,j}^{(k,l)}$ is the number of grid-based points belonging to both the $i$-th and $j$-th segments in the $k$-th and $l$-th frames, respectively. Also, the corresponding segment pairs, $\hat{\mathcal{P}}^{(k)}$, whose correlation score is the maximum, is defined as follows:
\begin{equation}
    \hat{\mathcal{P}}^{(k)} = \mathcal{P}^{(k,l_m)},
\end{equation}
where $l_m$ is
\begin{equation}
    l_m = \underset{l}{\textrm{argmax}} (score^{(k,l)}),
\end{equation}
a score between the $(k,l)$-th frames is
\begin{equation}
	score^{(k,l)} = \frac{\sum_{i=1}^{g}{\left( N_{i}^{(k)} \times \underset{j}{\max} {C}_{ij}^{(k,l)} \right)}}{\sum_{i=1}^{g}{N_{i}^{(k)}}},
\end{equation} 
and $g$ is the number of distinct motions. And then, we rearrange $\mathbf{G}$ to $\hat{\mathbf{G}}$ through $\hat{\mathcal{P}}^{(k)}$ as follows:
\begin{equation}
    \hat{G}^{(k)}_i \equiv 
    \hat{G}^{(k)}_{\hat{\mathcal{P}}^{(k)}_{i,\textrm{left}}} =
	\hat{G}^{(l_m)}_{\hat{\mathcal{P}}^{(k)}_{i,\textrm{right}}},
\end{equation}
so that we track the object that appeared previously or add a new segment on the unmatched pairs of the current frame. From now, we will call the value of $\hat{\mathbf{G}}$ as a label in the following paragraphs, while we have called the value of $\mathbf{G}$ as a segment.

\subsection{Dual-mode Motion Model}

Since the RGB-D camera has a limited depth range, it may fail to calculate scene flow vectors in some grid cells. Thus, we use a discrete statistical model which has a probability vector and an age as properties in order to track the static and dynamic parts in the image sequences persistently under the assumption that the static and dynamic elements do not appear or disappear abruptly on a frame by frame. The $i$-th element of the probability vector means the likelihood of corresponding pixel belonging to the $i$-th label. As shown in the right dashed box of \cref{fig:framework}, the algorithm first compensates the model through the previously estimated ego-motion to update it with the measurement corresponding to the identical 3D point. Then, we update the probability vector and age of the model based on certain criteria. Finally, labels can be selected as indices which indicate the maximum value in the probability vector. In this paper, each of $n$ grid cells has two models, i.e. apparent and candidate models. The apparent model indicates the estimated label currently, and the candidate model implies a hidden label which can appear later in the apparent model. The candidate model is designed to recognize again an object as static parts when the moving object stops.

\subsubsection{Model Update}

We denote the probability vector of a grid cell $i$ as $P_i^{(k)} \in \mathbb{R}^{n_{\textrm{obj}}\times 1}$ in the $k$-th frame, and the age of the grid cell in the $k$-th frame as $\alpha_i^{(k)}$, where $n_{\rm{obj}}=15$ is the designated maximum number of identified objects in a frame. Then, the probability vector and age are updated as follows:
\begin{align}
	P_i^{(k+1)} &= \frac{\tilde{\alpha}_i^{(k)}}{\tilde{\alpha}_i^{(k)}+1} \tilde{P}_i^{(k)} + \frac{1}{\tilde{\alpha}_i^{(k)}+1} \check{G}_i^{(k+1)}\\
	\alpha_i^{(k+1)} &= \tilde{\alpha}_i^{(k)} + 1,
\end{align}
where $\check{G}^{(k)} \in \mathbb{R}^{n_{\textrm{obj}}\times 1}$ is a binary-valued vector whose  $\hat{G}_i^{(k)}$-th element only is one and the others are zeros, and $\hat{G}_i^{(k)}$ is the result of segment matching algorithm in the $k$-th frame. $\tilde{P}_i^{(k)}$, $\tilde{\alpha}_i^{(k)}$ are the compensated parameters of the dual-mode motion model which will be discussed in \cref{sec:4:compensateSGM}. 

Contrary to \cite{moo2013detection} which measures pixel intensity to update models, we take a temporarily matched label, which is a result of segment matching, as measurements. However, our measurements might be incorrect, if scene flow may too fast to find an appropriate match and the properly-matched pair could not be found. Thus, we make some modification for the update of both models. The probability vector, ${}^{A}P$, and age, ${}^{A}\alpha$, of the apparent model are updated as follows: 
\begin{align}
	{}^{A}P_i^{(k+1)} &= \begin{cases}
    {}^{A}\tilde{P}_i^{(k)}, ~~~~~~~~~~~~~~~~~~~~~~~~ \text{if $\check{G}_i^{(k+1)} \neq \check{G}_i^{(k)}$}\vspace{3mm}\\
    \frac{1}{{}^{A}\tilde{\alpha}_i^{(k)}+1} ({}^{A}\tilde{\alpha}_i^{(k)} {}^{A}\tilde{P}_i^{(k)} + \check{G}_i^{(k+1)}), ~~ \text{otherwise}
    \end{cases}\nonumber\vspace{1mm}\\
	{}^{A}\alpha_i^{(k+1)} &= \begin{cases}
	{}^{A}\tilde{\alpha}_i^{(k)}, ~~~~~~~~~~~~~~~~~~~~~~~~ \text{if $\check{G}_i^{(k+1)} \neq \check{G}_i^{(k)}$}\vspace{2mm}\\bm
	\min{({}^{A}\tilde{\alpha}_i^{(k)} + 1, \alpha_{\rm{max}})}, ~~~~~~~~~~~~~ \text{otherwise}
	\end{cases}\nonumber
\end{align}
and the probability vector, ${}^{C}P$, and age ,${}^{C}\alpha$, of the candidate model are updated as follows:
\begin{align}
    {}^{C}P_i^{(k+1)} &= \begin{cases}
    {}^{C}\tilde{P}_i^{(k)}, ~~~~~~~~~~~~~~~~~~~~~~~~ \text{if $\check{G}_i^{(k+1)} = \check{G}_i^{(k)}$}\vspace{3mm}\\
    \frac{1}{{}^{C}\tilde{\alpha}_i^{(k)}+1} ({}^{C}\tilde{\alpha}_i^{(k)} {}^{C}\tilde{P}_i^{(k)} + \check{G}_i^{(k+1)}), ~~ \text{otherwise}
    \end{cases}\nonumber\vspace{1mm}\\
	{}^{C}\alpha_i^{(k+1)} &= \begin{cases}
	{}^{C}\tilde{\alpha}_i^{(k)}, ~~~~~~~~~~~~~~~~~~~~~~~~ \text{if $\check{G}_i^{(k+1)} = \check{G}_i^{(k)}$}\vspace{2mm}\\
	\min{({}^{C}\tilde{\alpha}_i^{(k)} + 1, \alpha_{\rm{max}})}, ~~~~~~~~~~~~~ \text{otherwise}\\
	\end{cases}\nonumber
\end{align}
Also, the candidate and apparent models are swapped if the age of the candidate model reaches the maximum age, $\alpha_{\rm{max}}$, or is larger than that of the corresponding apparent model. Because we treat a foreground object as a static element when the object stops. Thus, after the previously moving object stops, the corresponding apparent model is not updated whereas the candidate model is updated. Consequently, the age of only the candidate model increases and both models will be swapped with each other when the age of the candidate model becomes saturated or larger than that of the apparent model.

A label which is the output of the motion segment tracking is obtained from the probability vector of the apparent model with several criteria. The label is updated when the corresponding apparent model is initialized or updated. By doing so, we prevent the algorithm from prejudging the label of an unobserved or unmeasured grid cell while maintaining the previous label of the grid cell. Finally, the label is obtained from the indices that represent the maximum value in the probability vector of the apparent model.

\subsubsection{Compensate Model}\label{sec:4:compensateSGM}

In the previous section, we use the probability vector for updating and determining the label of a grid cell. These processes for the motion model assume that each grid cell represents a fixed point in the world coordinate consistently. However, in the case of a non-stationary camera, the result of the segment matching following the spatial segmentation cannot be used directly for updating the apparent and candidate model. In order to update the motion model, therefore, we compensate the model by warping.

\begin{figure*}[!t]
\begin{center}
\begin{tabular}{c c c c}
    \includegraphics[width=4.0cm]{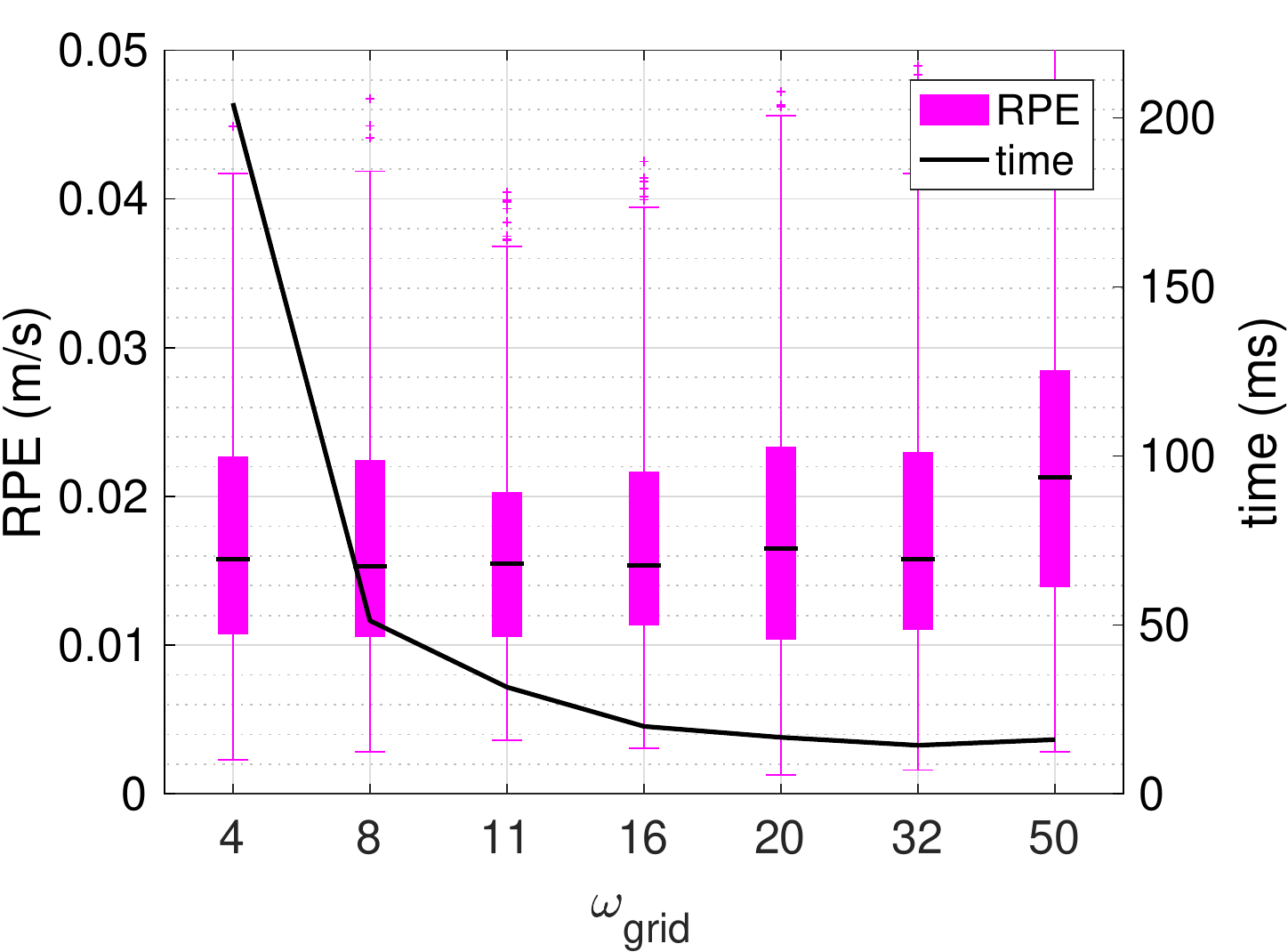} &
    \includegraphics[width=4.0cm]{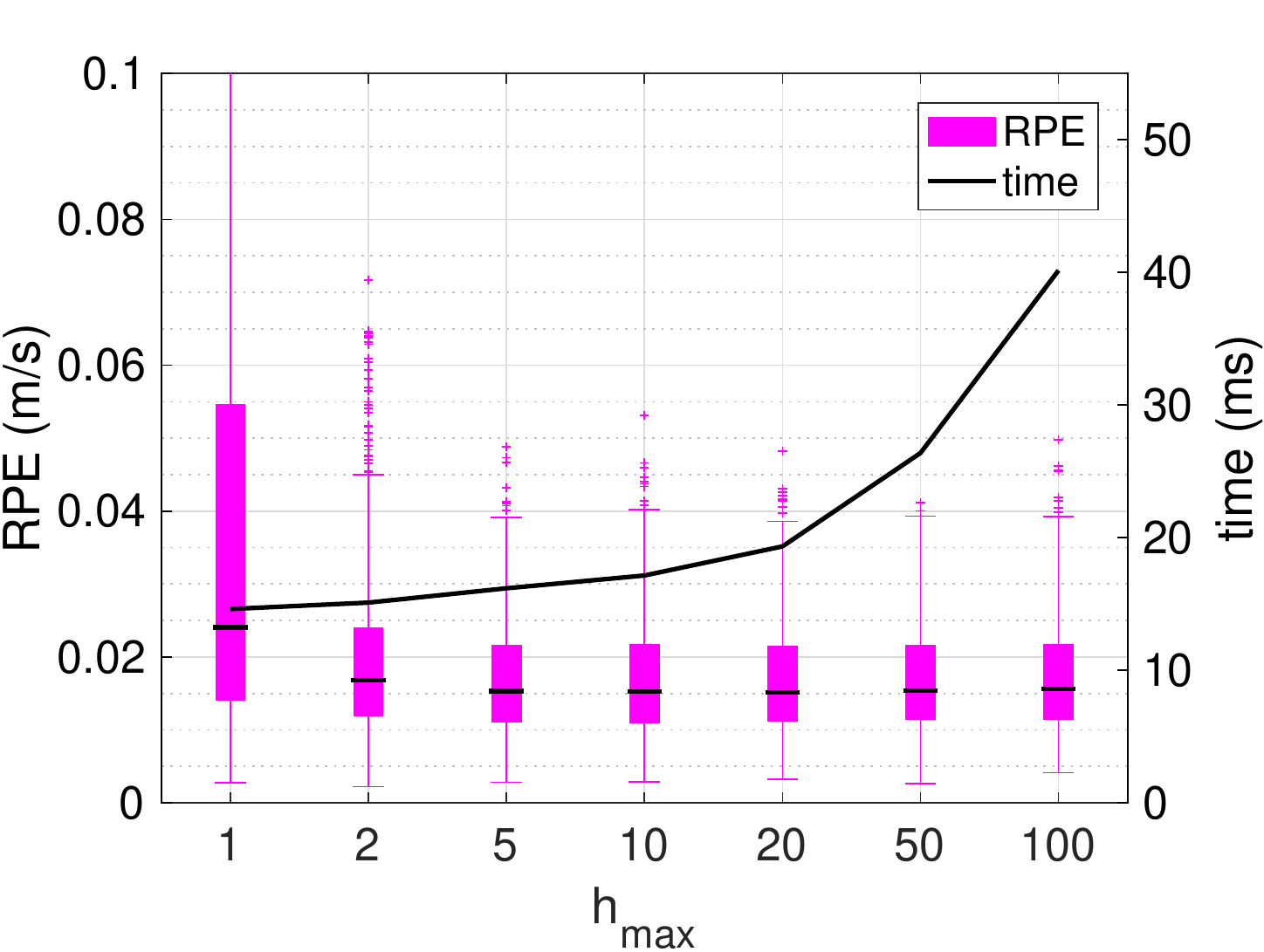} & 
    \includegraphics[width=4.0cm]{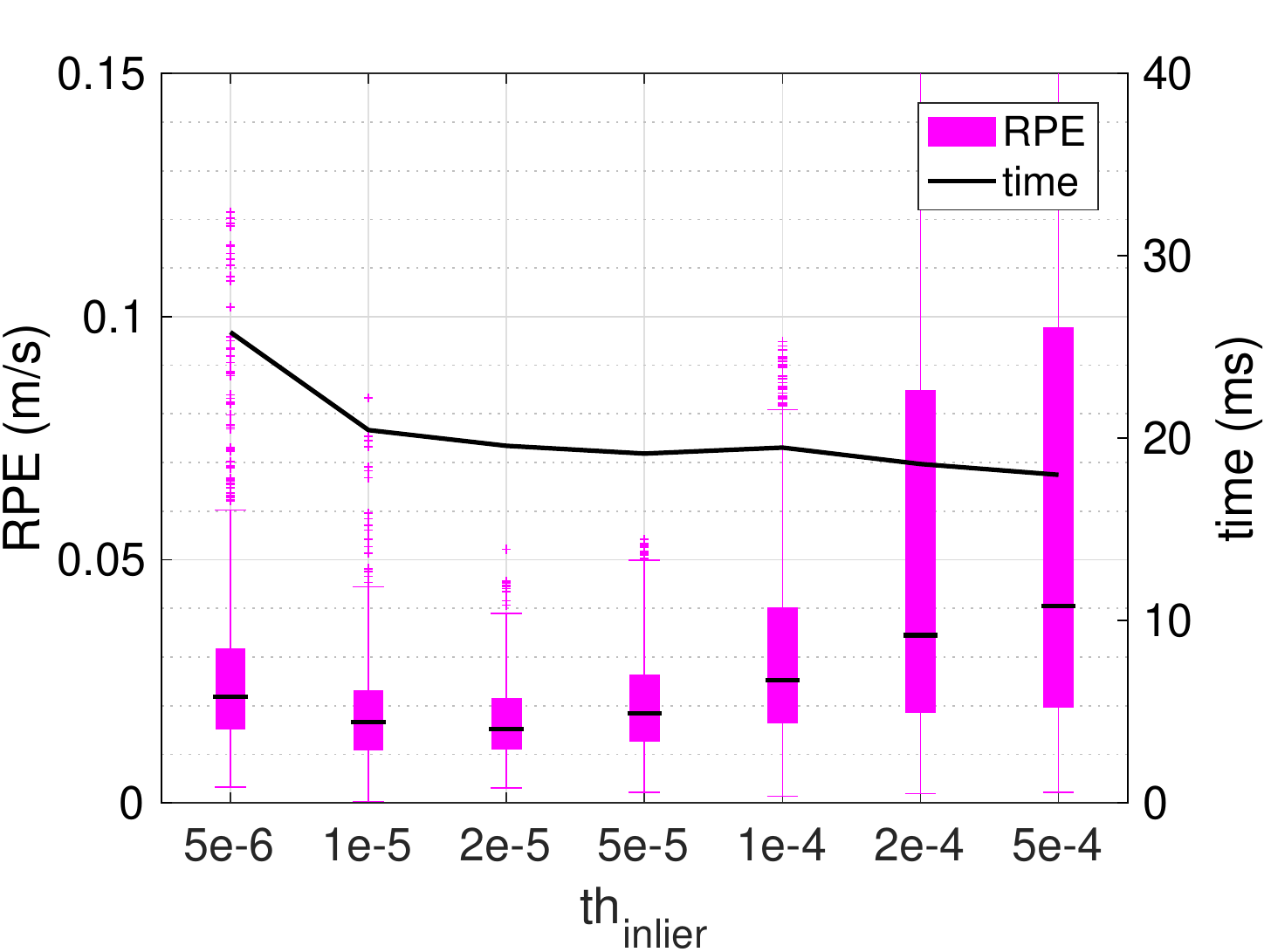} & 
    \includegraphics[width=4.0cm]{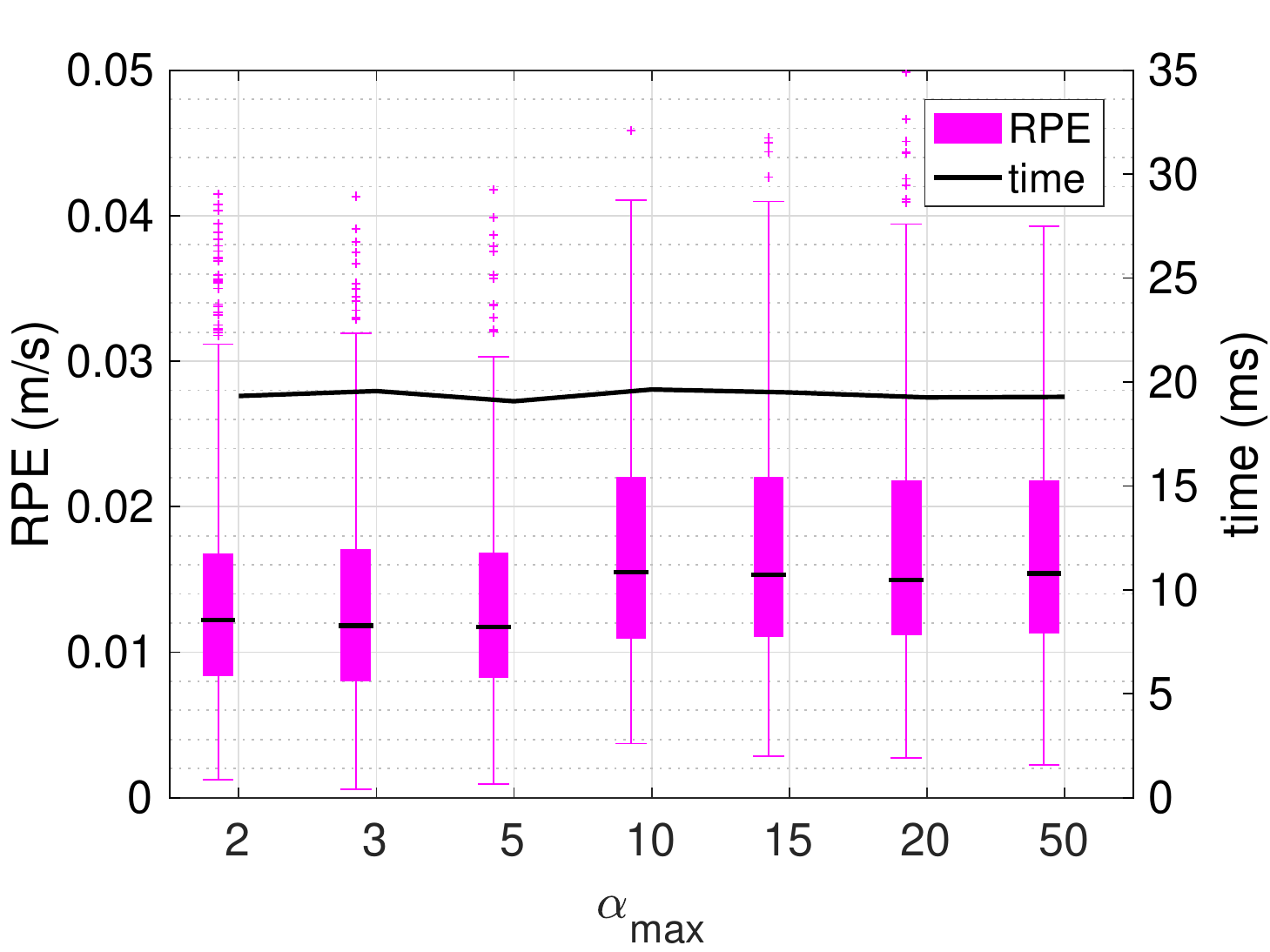} \\
\end{tabular}
\end{center}
    \vspace{-4mm}
    \caption{Variations of the parameters. The relative pose error are denoted as magenta boxplot with 3-$\sigma$ whiskers. Average computational time is represented as a black solid line with respect to the right y-axis.}
    \vspace{-3mm}
\label{fig:param}
\end{figure*}

Since each model has simple parameters such as probability vector and age, we use area-weighted interpolation. The current model is proportionally compensated with scene flow vectors in \cref{sec:2:sceneflow}. For grid cells that have valid scene flow vectors, these motion models are compensated individually on the two-dimensional image plane. We denote the set of grid cells overlapping with the current grid cell $i$ in the $(k+1)$-th frame as $\mathcal{A}_i^{(k+1)}$, weights for interpolation as $\omega_j$, and the region of overlap between the current grid cell $i$ and the previous corresponding grid cell as $R_j$ where $j$ is the index of element in $\mathcal{A}_i^{(k+1)}$. Then, the compensated probability vector and age are obtained as follows:
\begin{equation}
    \small{
	\tilde{P}_i^{(k)} = \sum_{j\in\mathcal{A}_i^{(k+1)}}{\omega_j P_j^{(k)}}, ~
	\tilde{\alpha}_i^{(k)} = \sum_{j\in\mathcal{A}_i^{(k+1)}}{\omega_j \alpha_j^{(k)}},}
\end{equation}
where $\omega_j$ are set to be proportional to $R_j$ and normalized to $\sum_{j}{\omega_j} = 1$.

\section{EVALUATION RESULTS}\label{sec:result}

This section is divided into three parts: analysis of parameters, description of the dataset, and quantitative evaluation of the odometry. In the validation, we used RGB-D images of 640$\times$480 pixels. Our algorithm is run on an Intel i7-7500U CPU at 2.7 GHz and Ubuntu 16.04 LTS.

\subsection{Parameters Analysis}

Here, we discuss the value of parameters introduced in the paper. These parameters are obtained from indoor environments including our dataset. Therefore, they need to be tuned for a new environment including outdoor environments.

\cref{fig:param} shows variations of the performance and computational loads according to parameters (from left): grid spacing, i.e. size of grid cell $w_{\rm{grid}}$, the maximum number of motion hypothesis $h_{\rm{max}}$, rigid transformation error threshold in motion hypothesis generation $th_{\rm{inlier}}$, the saturated age of motion model $\alpha_{\rm{max}}$. In each plot, one parameter varies while others remain the designated values given in the last paragraph of this subsection. As $w_{\rm{grid}}$ increases, the spatial density of the scene flow reduces, thus the computation time decreases reciprocally and Relative Pose Error (RPE) increases slightly. Therefore, this is advantageous in the presence of a small dynamic object. Also, if we choose a small value of $h_{\textrm{max}}$, it is difficult to extract all meaningful motions due to few tries in motion hypothesis searching, whereas a large value of $h_{\textrm{max}}$ can produce an erroneous label with a high computational load. Thus, if there are a lot of dynamic objects, it is recommended that the value of $h_{\textrm{max}}$ be not small. $th_{\rm{inlier}}$ is related to hypothesis searching; a low threshold makes the algorithm to extract motion hypothesis strictly. A small value of $\alpha_{\rm{max}}$ is vulnerable to erroneous label, whereas a high value makes the algorithm too insensitive not to notice that the previously moving object has stopped. 

From the above boxplot analysis, the parameters were set to: $w_{\rm{grid}}=16$, $h_{\rm{max}}=20$, $th_{\rm{inlier}}=3\times 10^{-5}$, $\alpha_{\rm{max}}=5$, which are well-tuned values for our indoor dataset. The parameters are kept the same over the dataset for consistency.

Rest of the parameters are set as follows:

\begin{itemize} 
    \item The number of points and the searching radius in \cref{sec:MHS} are set to $m = 7$ and $r_{\rm{search}} = 2$ pixels.
    \item The characteristic parameters and the minimum threshold of search weighting in \cref{sec:MHR} are set to $\lambda = 10^3$ and $\delta = 10^{-2}$, heuristically.
    \item DBSCAN parameters are set to $p_{\rm{min}}=1$, $\epsilon=0.005$.
\end{itemize}

\begin{figure}[t]
\begin{center}
    \includegraphics[width=8cm,height=4cm]{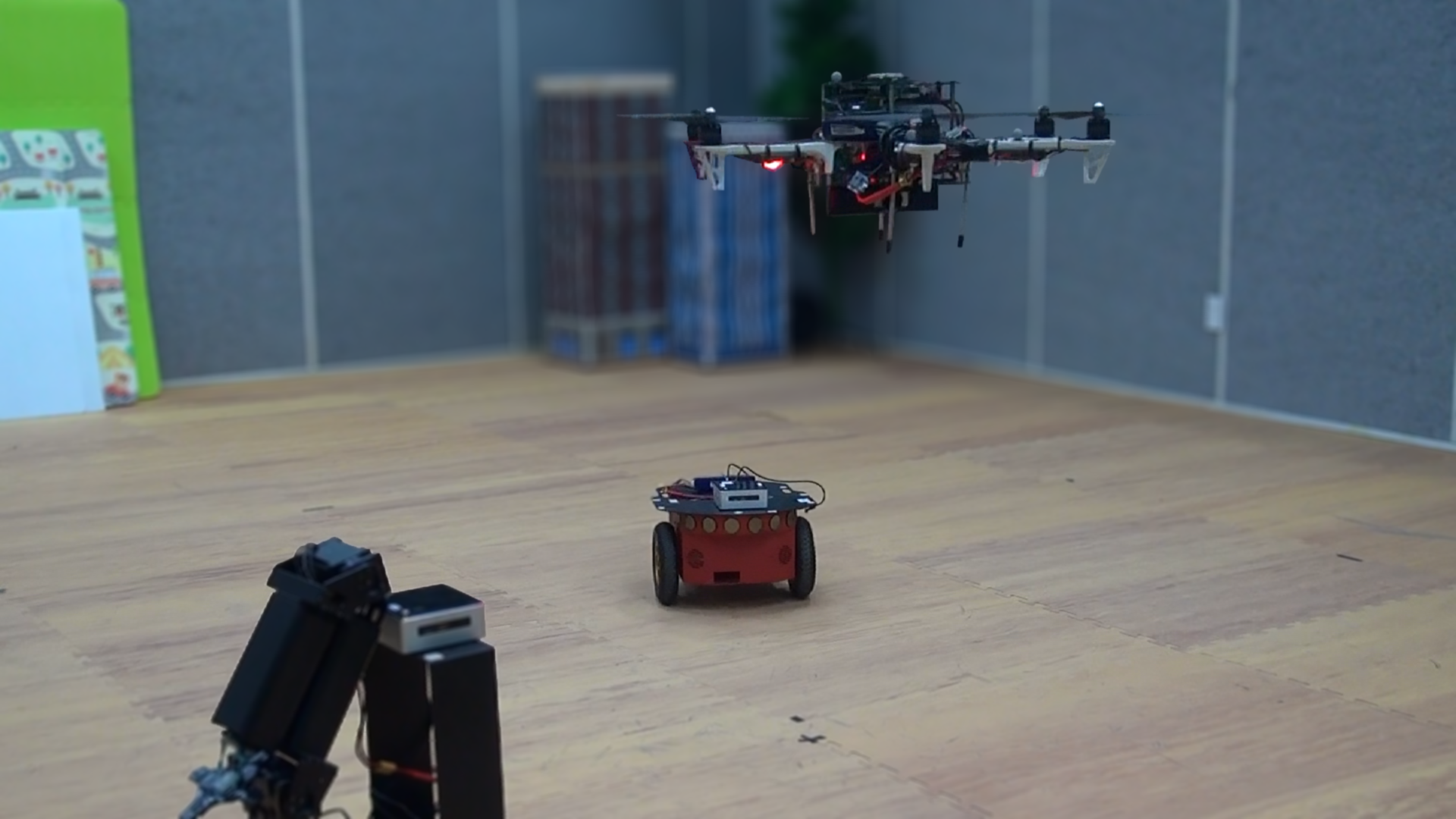}
    \caption{The validation environment for the \texttt{uav-flight-circular} sequence.}
    \label{fig:environment}
    \vspace{-8mm}
\end{center}
\end{figure}

\subsection{Dataset}

To the best of our knowledge, there is no well-known RGB-D dataset captured in a situation where a \textit{rigid moving object} appears. TUM dataset\cite{sturm2012benchmark} contains moving people but human body is not a rigid object and KITTI\cite{geiger2013vision} does not provide depth images. Therefore, in order to evaluate the proposed algorithm, we collected a dataset using ASUS Xtion RGB-D camera and Vicon motion capture system for ground truth (\cref{fig:environment}). Each sequence is composed of 16-bit depth images and 8-bit color images with a size of 640$\times$480. There are two kinds of sequences depending on whether they are captured from the static or non-stationary camera as shown in \cref{tbl:evaluation}. In the case of sequences using the static camera, we regard the origin of the world coordinate as the true position of the camera. On the other hand, for the non-stationary camera, we validate the performance of our visual odometry algorithm with Vicon measurements. The detailed description and download link of the dataset are available online at:
\begin{center}
    \small \texttt{http://sangillee.com/\_pages/icsl-de-dataset}
\end{center}

\begin{table*}[t]
\centering
\vspace{4mm}
\caption{Evaluation of visual odometry algorithms on our dataset.}\label{tbl:evaluation}
\begin{adjustbox}{width=.95\textwidth}
\vspace{4mm}
\begin{tabular}{c | c | cccccc } \noalign{\hrule height 1pt}
\multirow{2}{*}{Environment} & \multirow{2}{*}{Sequence} & \multicolumn{6}{c}{Relative Pose Error [m/s]} \\ \cline{3-8}
    & & Proposed & DVO & ORB-VO & BaMVO & Joint-VO-SF & StaticFusion\\ \hline \hline
    \multirow{3}{*}{\makecell{Static camera \& \\Dynamic environment}} & one-object-static & 0.0053 & {\color{myred} 0.6002} & \bf{0.0008} & 0.0017 & 0.0053 & {\color{myred} 0.2042}\\
    & two-object-static & 0.0221 & {\color{myred} 0.1673} & 0.0725 & \bf{0.0063} & 0.0172 & {\color{myred} 0.1603}\\
    & place-items & 0.0039 & 0.0550 & \bf{0.0017} & 0.0462 & 0.0344 & 0.0085\\ \hline
    \multirow{6}{*}{\makecell{Dynamic camera \& \\Dynamic environment}} & fast-object & \bf{0.0249} & {\color{myred} 0.4240} & {\color{myred} 0.3428} & {\color{myred} 0.2022} & 0.0405 & {\color{myred} 0.1726}\\
    & slow-object & \bf{0.0600} & {\color{myred} 0.1962} & {\color{myred} 0.1772} & {\color{myred} 0.1248} & 0.0724 & {\color{myred} 0.1311}\\
    & close-approach & \bf{0.0469} & {\color{myred} 0.2101} & 0.0931 & 0.0992 & 0.0707 & {\color{myred} 0.1360}\\
    & leading-pioneer & 0.0996 & {\color{myred} 0.1449} & \bf{0.0679} & {\color{myred} 0.2790} & {\color{myred} 0.1444} & {\color{myred} 0.5503}\\
    & uav-flight-static & \bf{0.0231} & {\color{myred} 0.3880} & {\color{myred} 0.2389} & {\color{myred} 0.2342} & 0.0279 & {\color{myred} 0.3812}\\ 
    & uav-flight-circular & \bf{0.0290} & {\color{myred} 0.2586} & {\color{myred} 0.1512} & {\color{myred} 0.4034} & 0.0737 & {\color{myred} 0.3109}\\ \hline \noalign{\hrule height 1pt}
\end{tabular}
\end{adjustbox}
\vspace{-2mm}
\end{table*}

\begin{table*}[t]
\centering
\caption{Evaluation of visual odometry algorithms on TUM dataset.}\label{tbl:evaluation_tum}
\begin{adjustbox}{width=.95\textwidth}
\begin{tabular}{c | c | cccccc } \noalign{\hrule height 1pt}
\multirow{2}{*}{Environment} & \multirow{2}{*}{Sequence} & \multicolumn{6}{c}{Relative Pose Error [m/s]} \\ \cline{3-8}
    & & Proposed & DVO & ORB-VO & BaMVO & Joint-VO-SF & StaticFusion\\ \hline \hline
    \multirow{4}{*}{\makecell{Dynamic camera \& \\Static environment}} & fr1/xyz & 0.0266 & {\color{myred} 0.1379} & \bf{0.0139} & {\color{myred}  0.1763} & 0.0174 & 0.0549\\
    & fr1/rpy & 0.0420 & 0.0406 & \bf{0.0303} & {\color{myred} 0.1858} & 0.0384 & 0.0889\\
    & fr1/desk & 0.0535 & 0.0675 & 0.0409 & {\color{myred} 0.2653} & \bf{0.0291} & {\color{myred} 0.1718}\\
    & fr1/floor & 0.0306 & {\color{myred} 0.1172} & \bf{0.0129} & {\color{myred} 0.1247} & 0.0266 & {\color{myred} 0.4150}\\ \hline
    \multirow{3}{*}{\makecell{Dynamic camera \& \\Dynamic environment}} & fr3/walking\_static & 0.0374 & {\color{myred} 0.2022} & {\color{myred} 0.1669} & 0.0939 & 0.0709 & \bf{0.0146}\\
    & fr3/walking\_xyz & {\color{myred} 0.2358} & {\color{myred} 0.2980} & {\color{myred} 0.2434} & {\color{myred} 0.1887} & {\color{myred} 0.2064} & \bf{0.0913} \\
    & fr3/sitting\_xyz & 0.0909 & 0.0367 & \bf{0.0110} & 0.0442 & 0.0444 & 0.0325\\ \hline \noalign{\hrule height 1pt}
\end{tabular}
\end{adjustbox}
\vspace{-2mm}
\end{table*}

\subsection{Visual Odometry}

For the quantitative comparison between the proposed algorithm and the current state-of-the-art visual odometry algorithms, we use RPE with a one second drift as proposed in \cite{sturm2012benchmark} since RPE is well-suited for evaluating the drift of visual odometry. Open-source algorithms were executed with default settings.

\cref{tbl:evaluation} and \cref{tbl:evaluation_tum} show the evaluation results for each sequence in our dataset and TUM dataset, respectively. The algorithm with the best result in each sequence is shown in bold, and the algorithm with an error greater than 0.1 m/s is in red. The proposed algorithm is compared with a well-known or state-of-the-art visual odometry algorithms such as DVO\cite{kerl2013dense}, ORB-VO\cite{mur2017orb}, BaMVO\cite{kim2016effective}, Joint-VO-SF\cite{jaimez2017fast}, and StaticFusion\cite{scona2018staticfusion}. Particularly, BaMVO, Joint-VO-SF, and StaticFusion were designed to be robust in dynamic environments likewise ours. In order to verify the performance as visual odometry, we evaluate a modified ORB-SLAM2, which is unable to detect and correct loop closure. We refer to this modified ORB-SLAM2 as ORB-VO. 

On the \texttt{one-object-static} sequence, algorithms show outstanding performance except for DVO. Because DVO is a direct dense method, it performs optimization across all pixels, thereby its performance is seriously influenced by dynamic elements. Next, when there appear two moving objects, the performance of our algorithm, BaMVO, and Joint-VO-SF is superior to the other algorithms. The performance of BaMVO is degraded when the object moves perpendicular to the principal axis in the \texttt{uav-flight-static, uav-flight-circular} sequences. ORB-VO tends to track features of a dynamic object when the object is observed for a long time in the \texttt{fast-object} sequence even if there are a lot of features on static backgrounds. Besides, Joint-VO-SF and StaticFusion fail optimization sometimes and does not perform well on some sequences. On the other hands, our algorithm shows a balanced and sufficient performance over all tested sequences. In the static environment sequences of TUM dataset, ORB-VO shows reliable performance superior to the other algorithms. Since the dynamic environment sequences of TUM dataset include the non-rigid human body, the proposed algorithm does not have an advantage over the existing algorithms, but it does not have a bad performance either. Please see the distribution of red texts for validating the reliability of the proposed algorithm.

\begin{table}[t]
\caption{Enhanced visual odometry algorithms for robustness.} \label{tbl:evaluation_adv}
\centering
\begin{adjustbox}{width=.45\textwidth}
\begin{tabular}{c | cc } \noalign{\hrule height 1pt}
\multirow{2}{*}{Environment} & \multicolumn{2}{c}{Relative Pose Error [m/s]} \\ \cline{2-3}
& \makecell{Proposed $\times$ ORB-VO} & ORB-VO \\ \hline \hline
    one-object-static & 0.0009 & \bf{0.0008} \\
    two-object-static & \bf{0.0242} & 0.0725 \\
    place-items & 0.0053 & \bf{0.0017} \\ \hline
    fast-object & \bf{0.0232} & \textcolor{myred}{0.3428} \\
    slow-object & \textcolor{myred}{\bf{0.1553}} & \textcolor{myred}{0.1772} \\
    close-approach & \bf{0.0728} & 0.0931 \\ 
    leading-pioneer & $\times$ & \bf{0.0679} \\ 
    uav-flight-static & \bf{0.0078} & \textcolor{myred}{0.2389} \\ 
    uav-flight-circular & \bf{0.0714} & \textcolor{myred}{0.1512} \\ \noalign{\hrule height 1pt}
\end{tabular}
\end{adjustbox}
\vspace{-6mm}
\end{table}

\begin{figure*}[!t]
\begin{center}
\begin{tabular}{c}
\vspace{2mm}
    \includegraphics[width=0.95\textwidth]{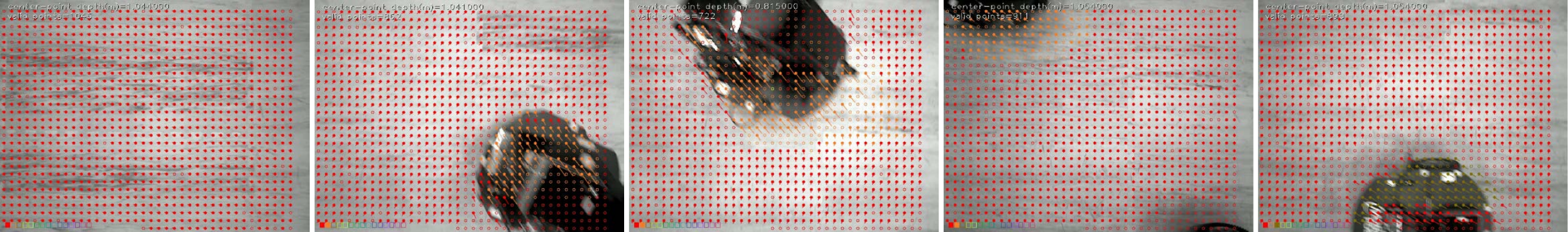} \\
    \includegraphics[width=0.95\textwidth]{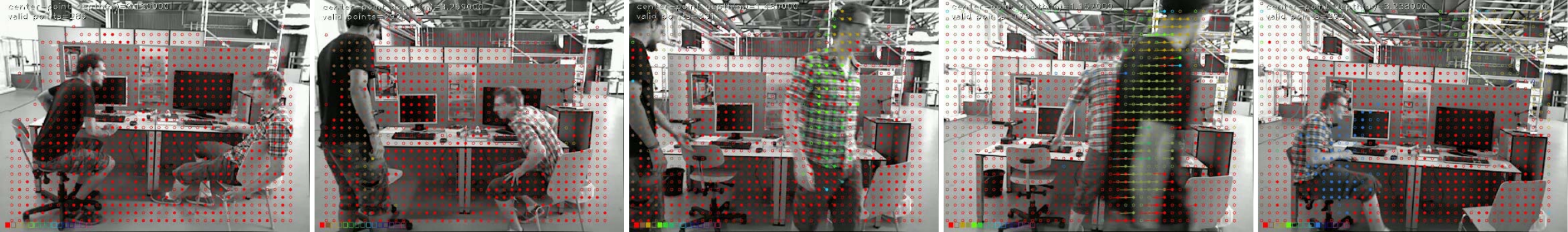}
\end{tabular}
\end{center}
    \vspace{-2mm}
    \caption{Some of the sequence images and the segmentation results that the proposed algorithm provides. Filled circles mean grid cell that has accurate scene flow and valid label. (Recommended to print out in color)}
    \vspace{-4mm}
\label{fig:seq2}
\end{figure*}

Moreover, since the motion segmentation and the estimation parts of our algorithm are not strongly coupled with each other, it is possible to incorporate the proposed motion segmentation into the existing visual odometry algorithms in order to improve their robustness in dynamic environments. As shown in \cref{tbl:evaluation_adv}, we validate the combination of ORB-VO and our motion segmentation. We can see that the proposed motion segmentation enhances the odometry performance compared with the original version. The \texttt{leading-pioneer} sequence contains only a small amount of valid depth for the entire time, so the feature points are not extracted enough causing failure. One way to improve the performance in this case is to convert from the grid-base to the dense method, and the accurate dense optical flow must be preceded first.

\subsection{Runtime Comparison}
The median runtimes of the compared algorithms are:

\vspace{-2mm}
\begin{itemize} 
\begin{multicols}{2}
    \item Proposed: 53 ms$^{1}$, \\ 267 ms$^{2}$
    \item DVO$^{2}$: 1.032 sec
    \item ORB-VO$^{1}$: 33 ms
    \item BaMVO$^{1}$: 278 ms
    \item Joint-VO-SF$^{1}$: 92 ms
    \item StaticFusion$^{1}$: 1.221 sec
\end{multicols}
\end{itemize}
\vspace{-2mm}

They are evaluated with $^{1}$ C++ implementation on a laptop computer as described in the first paragraph of \cref{sec:result} or $^{2}$ Matlab on a desktop computer (Intel i5-3770 at 3.4 GHz). Also, we make algorithms fetch RGB-D images of the same VGA size for fair comparison.

\section{CONCLUSIONS}\label{sec:disc}

In this paper, we proposed a real-time robust visual odometry algorithm via rigid motion segmentation using grid-based scene flow. The proposed algorithm is considerably more robust and accurate than the state-of-the-art visual odometry algorithms. For robustness, the proposed spatial motion segmentation uses scene flow to generate and search distinct motions with no prior information such as the shape or number of objects. Besides, temporal segmentation initializes and updates a dual-mode motion model of the grid cell so that our algorithm differentiates stationary background and dynamic objects robustly. Finally, the ego-motion is estimated by the use of scene flow vector fields belonging to the stationary background. An additional benefit of the proposed algorithm is that it can be combined with the existing visual odometry algorithms to improve their robustness in dynamic environments. Furthermore, the proposed approach can estimate the motion of moving objects unlike the other existing algorithms, so it can be employed as a part of an efficient dynamic obstacle avoidance algorithm for an autonomous robot by using the kinematic information of moving objects.


%%%%%%%%%%%%%%%%%%%%%%%%%%%%%%%%%%%%%%%%%%%%%%%%%%%%%%%%%%%%%%%%%%%%%%%%%%%%%%%%

%%%%%%%%%%%%%%%%%%%%%%%%%%%%%%%%%%%%%%%%%%%%%%%%%%%%%%%%%%%%%%%%%%%%%%%%%%%%%%%%

%%%%%%%%%%%%%%%%%%%%%%%%%%%%%%%%%%%%%%%%%%%%%%%%%%%%%%%%%%%%%%%%%%%%%%%%%%%%%%%%
%\section*{APPENDIX}
%\section*{ACKNOWLEDGMENT}
%
%The preferred spelling of the word �acknowledgment� in America is without an �e� after the �g�. Avoid the stilted expression, �One of us (R. B. G.) thanks . . .�  Instead, try �R. B. G. thanks�. Put sponsor acknowledgments in the unnumbered footnote on the first page.

%%%%%%%%%%%%%%%%%%%%%%%%%%%%%%%%%%%%%%%%%%%%%%%%%%%%%%%%%%%%%%%%%%%%%%%%%%%%%%%%

%References are important to the reader; therefore, each citation must be complete and correct. If at all possible, references should be commonly available publications.

\bibliographystyle{IEEEtran}
\bibliography{egbib}

\end{document}